\documentclass{article}

\PassOptionsToPackage{numbers, compress}{natbib}

\usepackage[preprint]{neurips_2023}

\usepackage[utf8]{inputenc} 
\usepackage[T1]{fontenc}    
\usepackage{hyperref}       
\usepackage{url}            
\usepackage{booktabs}       
\usepackage{amsfonts}       
\usepackage{nicefrac}       
\usepackage{microtype}      
\usepackage{xcolor}         
\usepackage[final]{graphicx}
\usepackage{subcaption}
\usepackage{multirow}
\usepackage{amsmath}
\usepackage{tikz}
\usepackage{listings,newtxtt}

\definecolor{deepblue}{rgb}{0,0,0.6}
\definecolor{deepred}{rgb}{0.6,0,0}
\definecolor{deepgreen}{rgb}{0,0.5,0}

\definecolor{xcodegray}{HTML}{707f8c}
\definecolor{xcodepurple}{HTML}{ad3da4}
\definecolor{xcodedarkpurple}{HTML}{703daa}
\definecolor{xcodered}{HTML}{d12f1b}

\lstdefinestyle{lststyle}{
language=Python,
basicstyle=\ttfamily\small,
commentstyle=\color{xcodegray},
otherkeywords={self},              
keywordstyle=\color{xcodepurple},
emph={flow_warping,SGD,ControlNet,cuda,cpu,temporal_consistency_optimization, depth2normal},  
emphstyle=\color{xcodedarkpurple}, 
stringstyle=\color{xcodered},
frame=tb,                          
showstringspaces=false 
}
\input{defs}

\title{Video ControlNet: Towards Temporally Consistent Synthetic-to-Real Video Translation Using \\ Conditional Image Diffusion Models}

\author{%
  Ernie Chu, Shuo-Yen Lin, Jun-Cheng Chen \\
  Research Center for Information Technology Innovation\\
  Academia Sinica\\
  Taipei, Taiwan \\
  \texttt{\{shchu,joseph,pullpull\}@citi.sinica.edu.tw} \\
}

\begin{document}

\begin{figure}[b!]
    \centering
    \includegraphics[width=1\linewidth]{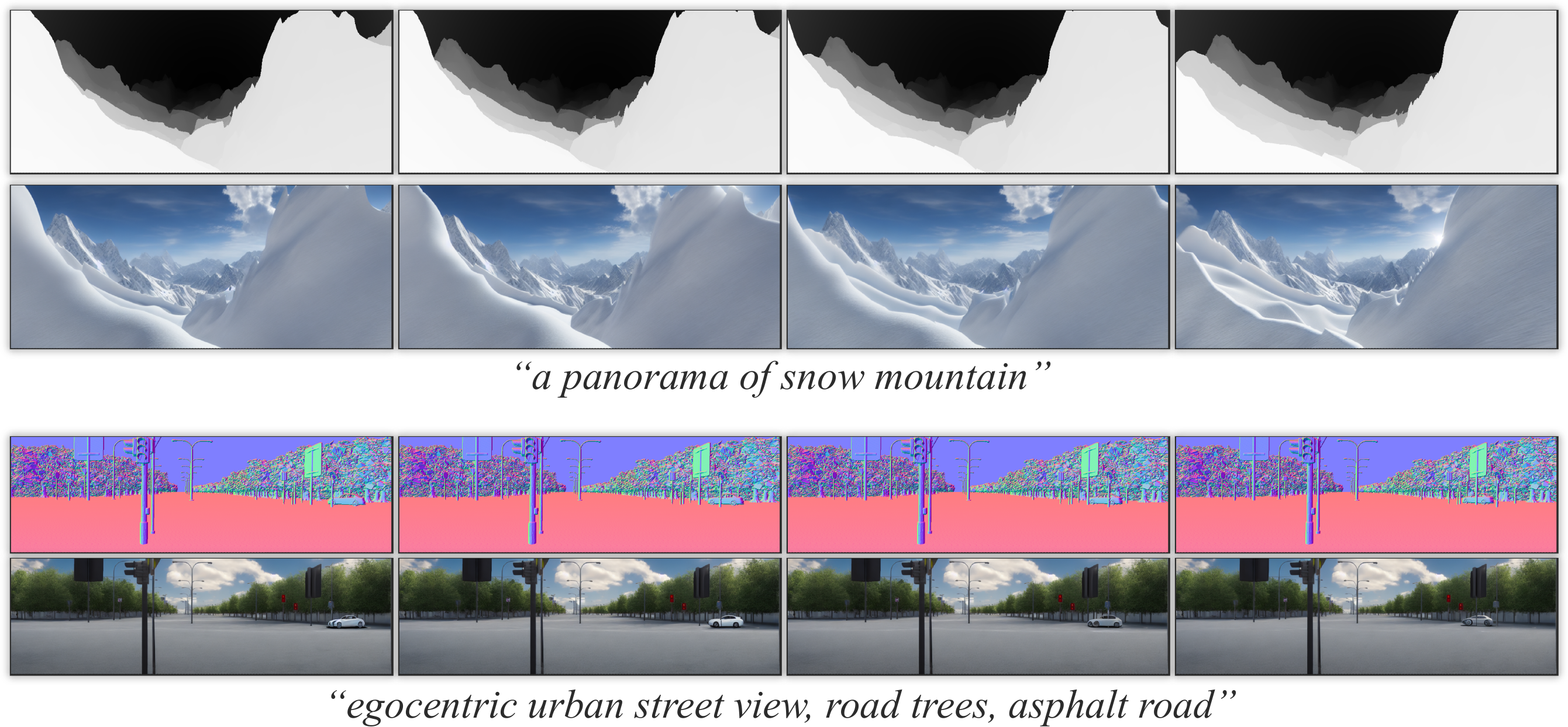}
    \caption{Video ControlNet samples based on synthetic assets from MPI Sintel and VKITTI.}
    \label{fig:my_label}
\end{figure}

\maketitle

\begin{abstract}
In this study, we present an efficient and effective approach for achieving temporally consistent synthetic-to-real video translation in videos of varying lengths. Our method leverages off-the-shelf conditional image diffusion models, allowing us to perform multiple synthetic-to-real image generations in parallel. By utilizing the available optical flow information from the synthetic videos, our approach seamlessly enforces temporal consistency among corresponding pixels across frames. This is achieved through joint noise optimization, effectively minimizing spatial and temporal discrepancies. To the best of our knowledge, our proposed method is the first to accomplish diverse and temporally consistent synthetic-to-real video translation using conditional image diffusion models. Furthermore, our approach does not require any training or fine-tuning of the diffusion models. Extensive experiments conducted on various benchmarks for synthetic-to-real video translation demonstrate the effectiveness of our approach, both quantitatively and qualitatively. Finally, we show that our method outperforms other baseline methods in terms of both temporal consistency and visual quality.

\end{abstract}

\section{Introduction}
Diffusion Models (DMs) \cite{ho2020ddpm,song2021score} have demonstrated unprecedented capability in high-dimensional density modeling, particularly in image generation \cite{
dhariwal2021diffusion,nichol2021improved,saharia2022palette}. The predictable scalability of DMs also gives the communities and big tech companies confidence to develop large text-to-image DMs \cite{saharia2022imagen,ramesh2022dalle2}, most notably the open-sourced Stable Diffusion \cite{rombach2022LDM}. Finetuing these large pre-trained DMs opens up numerous novel applications \cite{brooks2023instructpix2pix,ruiz2022dreambooth,zhang2023controlnet}. ControlNet \cite{zhang2023controlnet}, being an emergent star among these applications, provides a strong conditioning mechanism that controls the generated images by low-level conditions, such as per-pixel depths, surface normals, human poses, \etc. ControlNet also inherited the capability to interpret text prompts in order to control high-level semantics, textures, and colors, while preserving the scene and object structures.

ControlNet also has potential to replace traditional graphics rendering since depths and normals are common assets that can be easily exported from 3D modeling software, such as Blender \cite{blender}. Moreover, open source community has already developed ControlNet extensions for those artists who want to experience state-of-the-art realistic rendering. However, it is not a trivial task to extend ControlNet to video domain. Being an image-based generative model, ControlNet has no prior knowledge of how to maintain the temporal consistency when being applied to a series of animated conditions generated from the consecutive frames of a video.
The generated video often suffers from serious flickering and jumpy effects due to the inconsistency of textures and colors across frames. In this paper, we discuss the root cause of this behavior and propose an effective approach to mitigate this problem by utilizing the optical flow information from the synthetic video to enforce temporal consistency among the corresponding pixels across frames, unleashing the ControlNet's potential in synthetic-to-real video translation. It is worth noting that our approach does not require any training or fine-tuning ControlNet. With extensive experiments on various benchmarks for synthetic-to-real video translation, the results demonstrate the effectiveness of our approach, both quantitatively and qualitatively.

Our main contributions are summarized as follows:
\begin{itemize}
    \item We propose a simple and effective approach to perform temporally consistent synthetic-to-real video translation using the off-the-shelf ControlNet and the available optical flow information without requiring any further training and fine-tuning steps.
    \item We present quantitative measures to evaluate the temporal consistency of the synthesized videos based on optical flow information for the synthetic-to-real video translation task. 
    \item Furthermore, we also present and analyze different acceleration techniques for the optimization process to facilitate the proposed approach for a wider range of users and applications.
\end{itemize}

\section{Background}
In this section, we briefly review the two most pertinent previous works to the proposed approach: Stable Diffusion (Section \ref{sec:stable-diffusion}) and ControlNet (Section \ref{sec:control-net}).

\subsection{Stable Diffusion}
\label{sec:stable-diffusion}
Stable Diffusion \cite{rombach2022LDM} is an text-to-image Diffusion Model \cite{ramesh2022dalle2,saharia2022imagen} trained on a vast collection of image-text pairs in LAION \cite{schuhmann2022laion5b}. This model uses the cross-attention mechanism \cite{vaswani2017transformer} to incorporate semantic information into the underlying U-Net backbone, enabling the U-Net to effectively capture the conditional distribution of the observed images given the accompanying textual conditions. The textual condition is derived from the raw text strings through a pre-trained Large Language Model \cite{devlin2019bert,radford2021clip}.

\paragraph{Diffusion Models.} Stable Diffusion's base model, Latent Diffusion \cite{rombach2022LDM}, draws inspiration from the formulation introduced by Ho \etal\cite{ho2020ddpm} for Diffusion Models and utilizes the sampling scheduler proposed by Song \etal\cite{song2021ddim}. However, Latent Diffusion introduces a modification by confining the diffusion and denoising process to a VAE \cite{kingma14vae} latent space that has substantially lower dimensions compared to the observable space. This design choice enables notable acceleration in the overall process. Specifically, if we denote the encoded latent of an observed sample as $z^0$, the diffused latent is formally defined as
\begin{equation}
\label{diffusion}
z^{\ell} = \sqrt{\bar{\alpha}^{\ell}} z^0 + \sqrt{1-\bar{\alpha}^{\ell}} z^L, \hspace{1em}
\end{equation}
where $L$ is the total number of denoising steps, $\ell \in [0, L]$ is the index of the intermediate denosing step, $z^L \sim \mathcal{N}(0, I)$ and $z^{\ell}$ are the noise latents at the $L$-th and $\ell$-th step respectively, and $\bar{\alpha}^{\ell}$ is a strictly positive scalar-valued function of $\ell$ within $[0, 1]$. $z^L$ is also known as the initial noise. The U-Net backbone is trained to reconstruct $z^L$ from individual samples $z^{\ell}$ that span across various noise levels. As a result of this training, the model can accurately estimate the marginal distribution of $z^0$ in a recursive denoising process.

\paragraph{Implicit Diffusion Models.}
The sampling scheduler proposed in \cite{song2021ddim} reparameterizes the original denoising formula in \cite{ho2020ddpm} by isolating the noise term induced by the sampling process after each denoised distribution modeling. The scheduler further offers a scalar $\sigma^{\ell}$ on each isolated noise. When $\sigma^{\ell}$ is set to $0$ for all $\ell$, the entire diffusing and denoising process becomes deterministic, dubbed as DDIM. Denoising with non-Markovian noise levels under such setting comes with less quality degradation, thus enabling more efficient sampling. Song \etal \cite{song2021ddim} also found that for the same initial noises $z^L$ in DDIM, the samples generated with only 20 denoising steps are already  similar to the ones generated with 1,000 steps in terms of high-level features, indicating $z^L$ alone would be an informative encoding of the denoised image. Such property permits semantic interpolation on different $z^L$, which we will discuss later in our proposed method.

\subsection{ControlNet}
\label{sec:controlnet}
ControlNet \cite{zhang2023controlnet} is a trainable adapter designed to work in conjunction with Stable Diffusion, and it is initialized by partially copying the backbone U-Net of Stable Diffusion. The adapter plays a crucial role in preventing catastrophic forgetting when finetuning Stable Diffusion using additional image-text pairs and extra conditions. By incorporating the visual conditions through the adapter, ControlNet can effectively perform universal image-to-image translation while retaining the diverse, high-quality, and the promptable synthesis capability inherited from Stable Diffusion.

ControlNet's primary application lies in utilizing structural visual conditions, such as Canny edge map, per-pixel depths, normals, \etc, to exert control over the content and structure of synthesized images. Additionally, text prompts can be employed to specify the desired colors and textures for the synthesized results. 
This opens up possibilities for synthesizing video frames by utilizing a sequence of visual conditions, which can be inferred from real videos or extracted from animation assets. However, although the input visual conditions are temporally consistent and the content and structure in the generated video are somehow maintained, issues may still arise with regard to color and texture consistency. These inconsistencies may manifest as abrupt changes or jumps in colors and textures throughout the synthesized video frames.

There are two primary factors contributing to this cause.
\paragraph{The inconsistency comes from the entropy gap.}
First, the textual condition governing the textures is more abstract and operates at a higher level compared to the visual condition. Consequently, during training, a single text prompt often corresponds to multiple pixel combinations, forcing the model to rely on the difference in noise. In other words, the entropy present in the textual condition is significantly lower than that in the resulting images. Consequently, the model tends to rely on the random noise to bridge the entropy gap. Conversely, the visual condition typically possesses the same resolution and comparable entropy to the output image.

\paragraph{Altering the initial noise changes the semantic.}
Second, the initial noise samples used in denoising process significantly influence the synthesis of textures due to the aforementioned factor. Adjusting the noises for generating each frame based on the dynamics derived from the visual conditions is not a straightforward task, as spatially translating or warping the noises can alter the output semantics. To address this challenge, we propose a simple yet effective method in our work. We approach the problem by formulating the temporal consistency of the generated video as an optimization problem, which can be efficiently solved using gradient descent.

\label{sec:control-net}

\section{Video ControlNet}
\begin{figure}[t]
    \centering
    \includegraphics[width=1\linewidth]{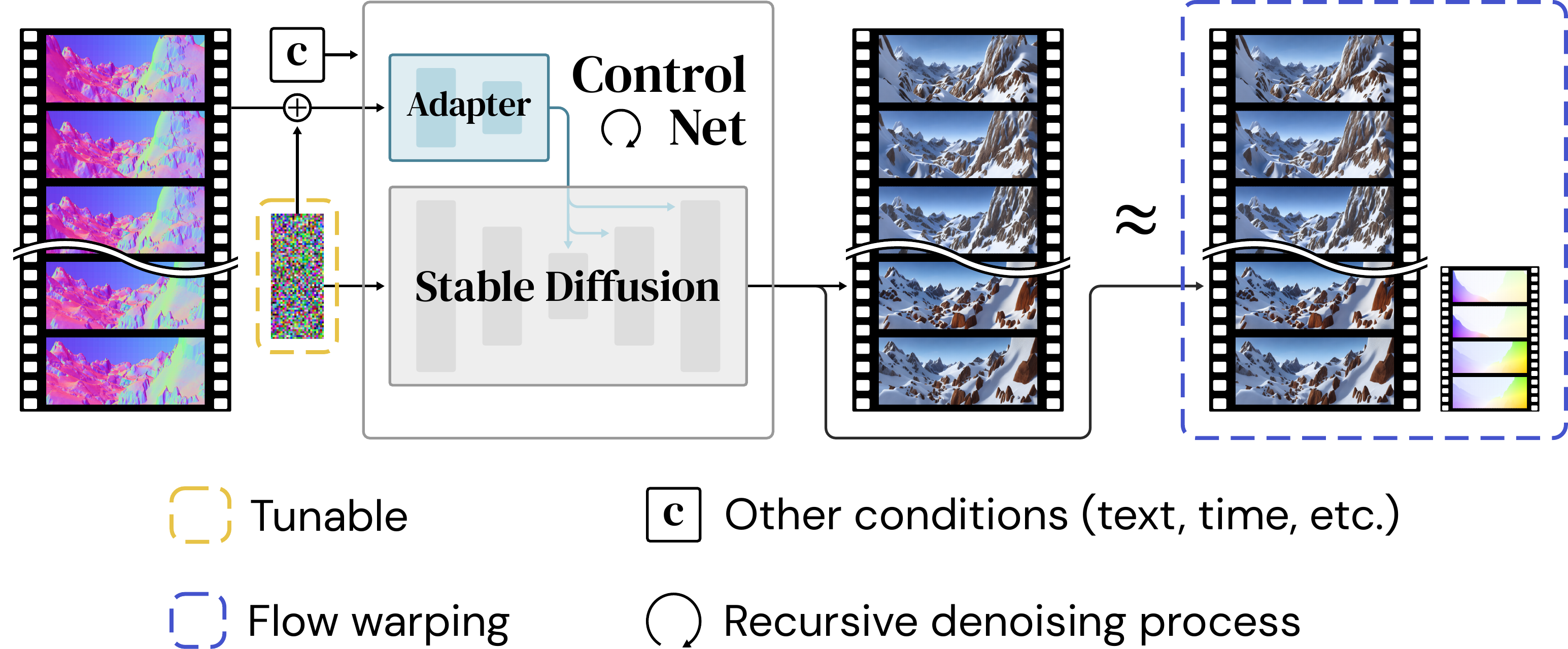}
    \caption{The high-level framework of the proposed Video ControlNet. We aim to minimize the distance between the output video and its flow-warped clone, enforcing the temporal consistency among the corresponding voxels. The gradient \textit{w.r.t.} the objective is back-propagated to the initial noise latent and the entire ControlNet is fixed.}
    \label{fig:arch}
\end{figure}

\label{sec:video-controlnet}

We target the main application of Video ControlNet at synthetic-to-real video translation, where the precise per-pixel depths, normals, occlusions, and optical flows can be easily exported from 3D modeling software such as Blender \cite{blender}. Our method also works for real-world videos with off-the-shelf estimators that infer less precise ingredients.

The key idea behind Video ControlNet is to craft an input noise combination that maximizes the temporal consistency of the generated video. First, we introduce a simple flow warping algorithm that quantifies the temporal consistency and permits least squares optimization (Section \ref{sec:flow-warping}). Next, we show how to craft the optimal input noise sequence in a sliding-window fashion for variable-length video synthesis (Section \ref{sec:optimization}). Finally, we present practical techniques to accelerate the optimization speed (Section \ref{sec:accelerate}), which helps democratize Video ContorlNet. The overall architecture for the proposed method is illustrated in Figure \ref{fig:arch}.

\subsection{Flow warping and masking}
\label{sec:flow-warping}
Optical flow describes the dense correspondences among voxels in the adjacent video frames. It can be obtained from 3D modeling software or inferred from real-world videos. We leverage the correspondence information to find the common voxels across video frames. We aim to minimize the discrepancy of each common voxel in different frames.

Formally, for an RGB video with $T$ frames, let $I(x, y, t) \in \mathbb{R}^3$ be the intensity of the voxel at location $(x, y, t)$. The flow warping function $\mathcal{F}$ that displaces the voxel from time $t+n$ to $t$ is given by
\begin{equation}
\label{eq:flow-warping}
\mathcal{F}_n: I(x, y, t) \mapsto I(x + \Delta x_{(t-n) \to t}, y + \Delta y_{(t-n) \to t}, t+n),
\end{equation}
where $(\Delta x_{(t-n) \to t}, \Delta y_{(t-n) \to t})$ are the optical flows of the voxel at $(x, y, t-n)$ pointing toward time $t$.

In addition to the optical flow, the occlusion map is critical to define the voxel correspondence as well. For voxels that do not present in the next frame, \ie occluded, their optical flows are undefined. Therefore, we should not associate them to any future voxels. To achieve this, a boolean mask that defines the region of interest is given by
\begin{equation}
\label{eq:masking}
\mathcal{M}(x, y, t) =
\begin{cases}
0, & \text{if } I(x, y, t+1) \text{ is occluded} \\
0, & \text{if } x + \Delta x_t \notin [0, W) \\
0, & \text{if } y + \Delta y_t \notin [0, H) \\
1, & \text{otherwise},
\end{cases}
\end{equation}
where $(W, H)$ is the frame width and height of the video.

\subsection{Temporal consistency optimization}
\label{sec:optimization}

Given a conditional distribution modeled by a trained ControlNet $\mathcal{CN}$, naively sampling a video from the distribution gives consistent visual structure across frames as long as the visual conditions (e.g. depth maps, normal maps, etc.) have the consistent structure. However, the textures and colors of each frame are conditioned on semantic conditions (e.g. text prompt), which are not specific enough to maintain the consistency of the corresponding voxel values across frames.

Given the aforementioned flow warping and masking, we can explicitly define the degree of consistency as follows. For simplicity, we use $I_t$ and $\mathcal{M}_t$ to denote $I(x, y, t)$ and $\mathcal{M}(x, y, t)$ for all $x \in [0, W)$ and $y \in [0, H)$, respectively. Let $I_t = \mathcal{CN}(z_t^L, c)$ be a video frame sampled from the ControlNet, where $c$ is the multi-modal condition and 
$z_t^L \sim \mathcal{N}(0, I)$ is the initial noise latent for the $t^{th}$ frame. To optimize the video for the best consistency, we define the discrepancy $\mathcal{D}(t)$ in a sliding window for the voxels between the $t^{th}$ frame $I_t$ and other $S$ frames prior to $I_t$ as
\begin{equation}
\mathcal{D}(t) = \sum_{s=0}^n \|\mathcal{M}_{t-s, t} \otimes \left( I_{t-s} - \mathcal{F}_{(s+1)}\left( I_{t-s}\right)\right)\|_2, \hspace{2em} 
\end{equation}
\begin{equation}
\mathcal{M}_{i,n} = \prod_{j=i}^n\mathcal{M}_{j} = \mathcal{M}_{i} \otimes \mathcal{M}_{i+1} \otimes \cdots \otimes \mathcal{M}_{n},
\end{equation}
where $S$ is the window size that determines how many previous frames are to be considered when optimizing for consistency, $n = \min(S, t)$, and $\otimes$ denotes Hadamard product. Since increasing $S$ introduces nearly zero overhead, we assume $S = T$ unless otherwise stated.

\paragraph{Crafting optimal noise.}
While finetuning the ControlNet, the objective may result in a degenerate solution where the output video collapse to a constant. We thus opt for optimizing the input noises $z^L = \{z^L_t \mid t \in [0, T)\}$ and keep ControlNet fixed. That is to say, an optimal noise for generating temporally consistent video should minimize the summation over normalized discrepancy $\mathcal{D}$ at each timestamp, such as
\begin{equation}
\label{eq:opt-noise}
z_*^L
= \argmin_{
z^L
}\sum_{t=0}^{T-1}
\mathcal{D}(t) / \left(\min(S, t) + 1\right),
\end{equation}
where $z_*^L$ is the optimized initial noise latent for video synthesis after iterative optimization. As discussed in the end of Section \ref{sec:controlnet}, the input noise dominantly determines the final textures and colors. We use a common noise across frames, \ie $z_0^L = z_1^L = \cdots = z_{T-1}^L$, to reduce the drastic change of textures. This provides significantly smaller discrepancy at the start of the optimization and speeds up the entire process.

\subsection{Accelerating the optimization process}
\label{sec:accelerate}
To improve the efficiency for the whole optimization process, we propose two acceleration techniques.

\paragraph{Optimizing different levels of noise.} In the previous section, we optimize the initial noise $z^L$ with a noise level $L$, which corresponds to a pure noise sampled from standard normal distribution. We can instead target at other noise levels $\{\gamma \mid \gamma \in [1, L] \cap \mathbb{Z} \}$ that are smaller than $L$. Since the number of model forward passes required to fully denoise the video is proportional to the noise level under the same sampling quality, optimizing the sample with smaller noise level directly accelerates the entire crafting process.

Concretely, we first perform a full denoising process on $z^L$ to obtain a clean latent $z^0$. Next, we reuse $z^L$ to diffuse the clean latent to a specific noise level $\gamma$ according to the diffusion process \eqref{diffusion} and obtain the optimal noisy latent $z^{\gamma}_*$ according to Eq. \eqref{eq:opt-noise}. 
Note that the textures and colors of a video may have been established when $\gamma$ is too small and thus impede the optimization.

\paragraph{Frame interpolation.} As Song \etal \cite{song2021ddim} have discovered, the high level features of the DDIM sample is encoded by $z^L$. So we can directly perform meaningful interpolation on the noisy latents $z^L$. Therefore, when computation resources are intensive, we can optimize every $k$ keyframe only and interpolate the latents $\{z^L_t \mid t \equiv 0 \pmod k\}$ of keyframes to obtain in-between latents post-hoc. To ensure the number of frames in the output video is invariant to $k$, we define the number of residual frames as $(T-1) \mod k$. The residual frames are optimized together with the keyframes. Formally, each noisy latent of the $t^{th}$ frame is given as
\begin{equation}
z^L_t = 
\begin{cases}
z^L_t, & t \geq T - \left((T-1) \mod k\right) \\
z^L_t, & t \equiv 0 \pmod k \\
Slerp \left( z^L_{\lfloor t \rfloor_k}, z^L_{\lceil t \rceil_k}; \frac{\left| t - \lceil t \rceil_k \right|}{k} \right), & \text{otherwise},
\end{cases}
\end{equation}

where $Slerp \left(u, v; \alpha \right) =  \frac{\sin((1 - \alpha) \theta)}{\sin(\theta)} u + \frac{\sin(\alpha \theta)}{\sin(\theta)} v$ is the spherical linear interpolation \cite{shoemake1985slerp}, $\theta = \arccos\left(\frac{u \cdot v}{\norm{u}\norm{v}}\right)$ is the angle between $u$ and $v$, $\lfloor t \rfloor_k = k \left\lfloor \frac{t}{k} \right\rfloor$, and $
\lceil t \rceil_k = k \left\lceil \frac{t}{k} \right\rceil$. 

\section{Experiment}
In this section, we assess the proposed method in three aspects. First, to evaluate the pixel-level temporal consistency (Section \ref{sec:exp-px-temporal-consistency}), we employ an off-the-shelf optical flow estimator to infer optical flows from the generated video and compare the results with the ground truth flows. Since the estimator is trained with temporal consistent videos, it should generalize worse to videos with large voxel discrepancies. Second, to validate the instance-level temporal consistency and realness of the generated video (Section \ref{sec:exp-instance-temporal-consistency}), we employ an off-the-shelf object tracker. The model is trained with real driving footage and should generalize worse to video frames that are not consistent and realistic enough. Finally, we provide additional results on validating the proposed acceleration techniques (Section \ref{sec:exp-accelerate}).

\begin{figure}
\centering
\begin{subfigure}[b]{1\textwidth}
   \includegraphics[width=1\linewidth]{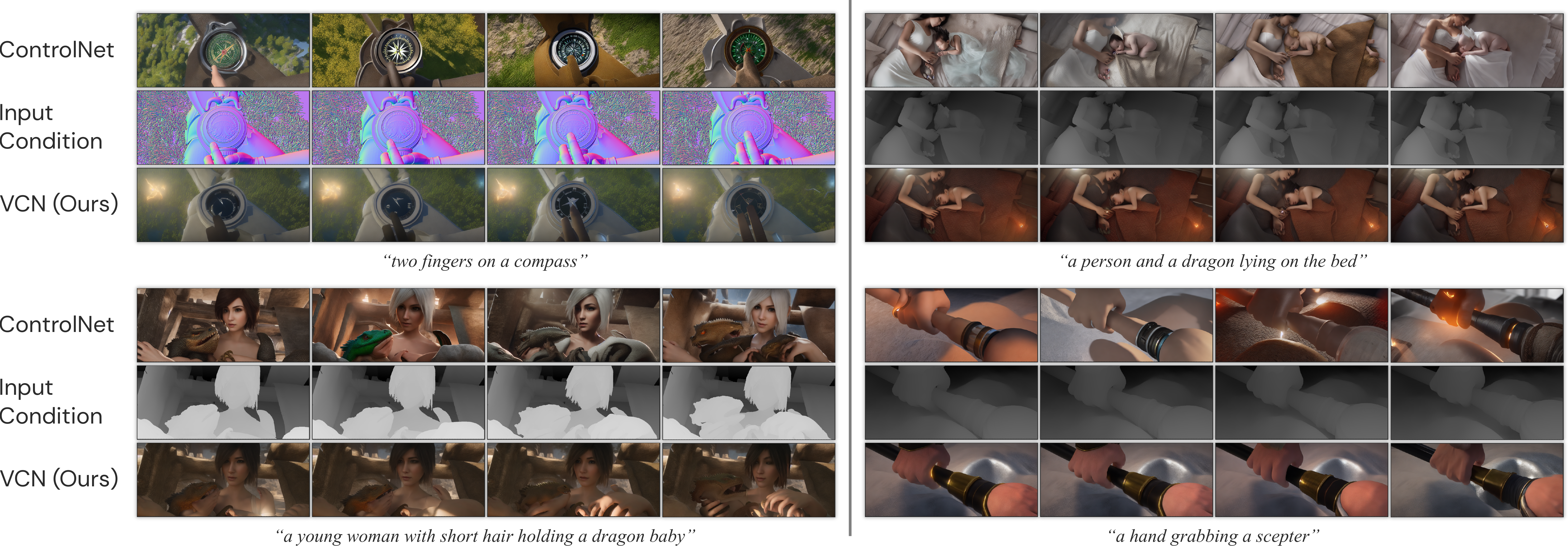}
\end{subfigure}

\begin{subfigure}[b]{1\textwidth}
    \includegraphics[width=1\linewidth]{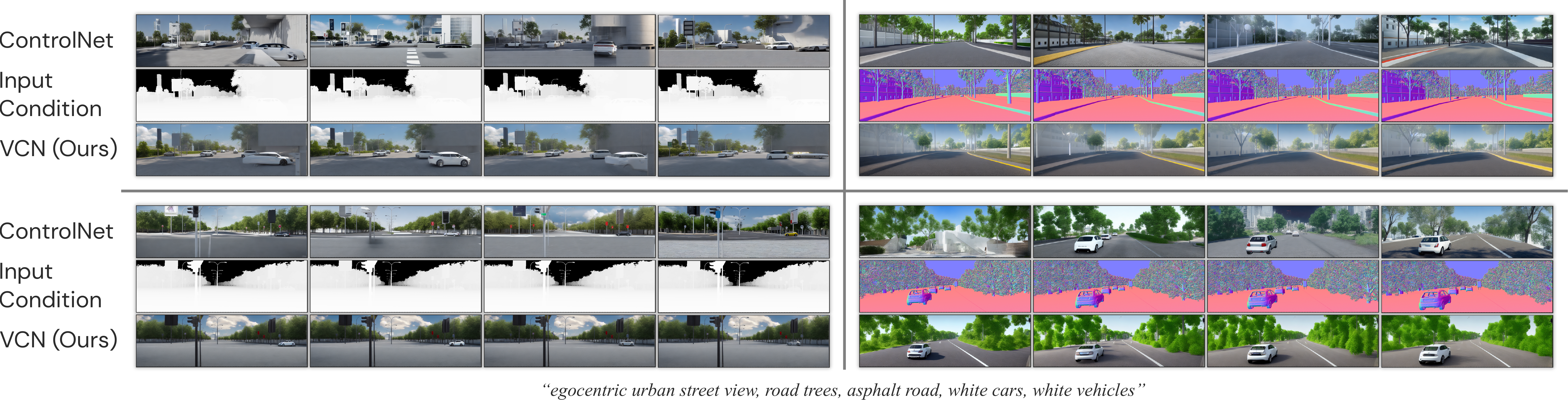}
\end{subfigure}
\caption{Video generated by ControlNet \cite{zhang2023controlnet} and the proposed Video ControlNet. The in-between condition images demonstrate the per-pixel depth map and normal map that are sent into the ControlNet adapter. \textit{Top:} Synthetic-to-real video translation on Sintel \cite{butler2012sintel}, \textit{Bottom:} on VKITTI \cite{gaidon2016vkitti}.}
\label{fig:qualitative}
\end{figure}

For all our experiments, we set the number of denoising steps to $10$ and $\eta$ in DDIM \cite{song2021ddim} to $0$ to save computation time. All videos are optimized by Adam optimizer \cite{kingma2013adam} with a learning rate of $10^{-3}$ to convergence. We run the experiment on 8 GeForce RTX 3080 24GB GPU for approximately 80 hours. In the main experiment, no acceleration technique in Section \ref{sec:accelerate} is used. That is to say, we optimize every frame of $z^T$. A visual comparison of the video samples of ControlNet \cite{zhang2023controlnet} and ours can be found in Figure \ref{fig:qualitative}.

\subsection{Pixel-level temporal consistency}
\label{sec:exp-px-temporal-consistency}

For this experiment, we use GMFlow \cite{xu2022gmflow} as the optical flow estimator and run it on the processed videos of the two datasets by the proposed approach for evaluation: MPI Sintel \cite{butler2012sintel} and Virtual KITTI \cite{gaidon2016vkitti} \footnote{The weight for KITTI is loaded in GMFlow when evaluating the Virtual KITTI part, so we can expect GMFlow may not generalize perfectly on Virtual KITTI.}. Both datasets are photo-realistic computer generated animations providing precise depths, optical flows, object trajectory, \etc, exported directly from the 3D modeling software. We use the depth maps and normal maps (derived from depth) as inputs to the Video ControlNet, respectively, and use the optical flows and occlusions (derived from optical flows when unavailable) to perform temporal consistency optimization.

We compare our method with two baselines: the original animation and vanilla ControlNet. Video frames generated by each method would be used by GMFlow to infer their optical flows. An average endpoint error (EPE) which calculates the Euclidean distance between the estimated flow and the ground truth flow would indicate the temporal consistency in each video clip, as reported in Table \ref{tab:flow-test}. Experimental results show that the EPEs are significantly lower in the videos generated by our method than the ones generated by ControlNet. The gap between the EPEs of ours and the original Sintel animations may be attributed to the fact that GMFlow is trained on the same split of the Sintel videos.

\begin{table}
  \caption{Average endpoint error (EPE) on the videos generated by the baseline methods and our Video ControlNet (VCN). The error are calculated by GMFlow's \cite{xu2022gmflow} predictions and the ground-truth optical flows. The best results are in \textbf{bold}.}
  \label{tab:flow-test}
  \centering
  \begin{tabular}{=l+l+c+c+c+c+c+c}
    \toprule
    Method & Condition & \multicolumn{5}{c}{MPI Sintel \cite{butler2012sintel}} \\
    \cmidrule(r){3-7}
    && Alley & Ambush & Bamboo & Bandage & Cave & Avg. \\
    \midrule
    ControlNet \cite{zhang2023controlnet} & Depth  & 71.73 & 111.82 & 39.52 & 73.34 & 72.94 & 73.87 \\
               & Normal & 68.84 & 124.05 & 98.23 & 85.00 & 93.07 & 93.84 \\
    VCN (Ours) & Depth  &  4.52 &  \textbf{28.68} &  \textbf{1.87} &  5.20 & 28.84 & 13.82 \\
               & Normal &  \textbf{4.51} &  30.75 &  3.81 &  \textbf{2.96} & \textbf{26.18} & \textbf{13.64} \\
    \rowstyle{\color{gray}}
    Animation  & N/A & 0.10 & 0.95 & 0.19 & 0.14 & 0.76 & 0.43 \\
    \midrule
    && \multicolumn{5}{c}{MPI Sintel \cite{butler2012sintel}} \\
    \cmidrule(r){3-7}
    && Market & Mountain & Shaman & Sleeping & Temple & Avg. \\
    \midrule
    ControlNet \cite{zhang2023controlnet} & Depth  & 137.66 & 80.10 & 75.57 & 101.72 & 103.17 & 99.64 \\ 
               & Normal &  68.84 & 124.05 & 98.23 & 85.00 &  93.07 & 93.84 \\
    VCN (Ours) & Depth  &  37.34 &  4.96 &  \textbf{2.44} &  6.40 &  \textbf{21.85} & 14.60 \\
               & Normal &  \textbf{18.85} &  \textbf{4.91} &  2.62 &  \textbf{2.87} &  22.61 & \textbf{10.37} \\
    \rowstyle{\color{gray}}
    Animation  & N/A    &   0.60 &  0.08 &  0.11 &  0.06 &   0.59 &  0.29 \\
    \midrule
    && \multicolumn{5}{c}{Virtual KITTI \cite{gaidon2016vkitti}} \\
    \cmidrule(r){3-7}
    && 0001 & 0002 & 0006 & 0018 & 0020 & Avg. \\
    \midrule
    ControlNet \cite{zhang2023controlnet} & Depth  & 48.26 & 38.21 & 40.40 & 57.53 & 52.23 & 47.33 \\
               & Normal & 52.61 & 47.26 & 42.97 & 58.51 & 55.53 & 51.38 \\
    VCN (Ours) & Depth  & \textbf{31.86} & \textbf{14.56} & \textbf{10.93} & \textbf{23.67} & \textbf{25.87} & \textbf{21.38} \\
               & Normal & 33.64 & 14.90 & 13.14 & 24.08 & 28.55 & 22.86 \\
    \rowstyle{\color{gray}}
    Animation  & N/A    & 36.32 & 16.30 & 11.80 & 30.59 & 33.46 & 25.69 \\
    \bottomrule
  \end{tabular}
\end{table}

\subsection{Instance-level temporal consistency and realness}
\label{sec:exp-instance-temporal-consistency}
To validate instance-level consistency and the potential of Video ControlNet in helping realistic rendering, we use CenterTrack \cite{zhou2020centertrack} to perform multi-object tracking. The experimental results in Table \ref{tab:mot-test} show that there are fewer ID switches and lower fragmentation in our generated videos compared to the ones of vanilla ControlNet. However, some MOT metrics are better on the videos sampled from ControlNet. We conjecture that this is because the proposed optimization sometimes leads to slightly blurry videos in order to attain a higher temporal consistency.

\begin{table}
  \caption{Multi-object tracking results on the videos generated by the baseline methods and our Video ControlNet (VCN). We report the CLEAR MOT metric \cite{bernardin2014clear}, including MOT Accuracy (MOTA), MOT Precision (MOTP), Mostly Tracked ratio (MT), Mostly Lost ratio (ML), ID switches (I), Fragmentation (F), and the detector's preicision (P) and recall (R). All numbers are calculated by the predictions of CenterTrack \cite{zhou2020centertrack}. Best results are in \textbf{bold}.}
  \label{tab:mot-test}
  \centering
  \begin{tabular}{=l+c+c+c+c+r+r+c+c}
    \toprule
    Method & MOTA $\uparrow$ & MOTP $\uparrow$ & MT $\uparrow$ & ML $\uparrow$ & I $\downarrow$ & F $\downarrow$ & P $\uparrow$ & R $\uparrow$ \\ 
    \midrule
    ControlNet-D & 31.07 & 72.38 & 12.08 & 42.03 & 1,000 & 1,192 & 89.71 & \textbf{55.78} \\ 
    ControlNet-N & 30.44 & \textbf{81.71} & \textbf{13.53} & 36.72 &  476 &  685 & 93.93 & 47.63 \\ 
    VCN-D (Ours)       & \textbf{42.48} & 69.31 &  9.18 & \textbf{52.66} &   59 &  260 & 89.12 & 51.43 \\ 
    VCN-N (Ours)      & 29.48 & 81.46 &  8.21 & 49.28 &   \textbf{31} &  \textbf{212} & \textbf{95.95} & 38.30 \\ 
    \rowstyle{\color{gray}}
    \rowstyle{\color{gray}}
    Real video       & 84.32 & 90.34 & 98.72 &  0.00 &    9 &   32 & 88.91 & 99.16 \\ 
    \bottomrule
  \end{tabular}
\end{table}

\subsection{Acceleration techniques}
\label{sec:exp-accelerate}
Besides the main experiments, we also conduct auxiliary evaluations to demonstrate how much time the proposed acceleration techniques (Section \ref{sec:accelerate}) save. Considering the optimizing noise level $\gamma$, we sweep four different possible values of $\gamma$ from noisy to clean. Concretely, we set the number of denoising steps to 10, \ie $z^{10} \sim \mathcal{N}(0, I)$, and tune $z^{\gamma}$ to synthesize temporally consistent videos, where $\gamma = \{2, 4, 7, 10\}$. On the other hand, we also run the optimization every $k$ keyframe only, where $k = \{1, 2, 3 ,4\}$. The results in Table \ref{tab:accelerate} show that the proposed acceleration techniques save proportional amounts of time without hurting the temporal consistency.

\begin{table}
  \caption{Average endpoint error (EPE) on the videos generated by our Video ControlNet with different accelerating hyperparameters. The error are calculated by GMFlow's \cite{xu2022gmflow} predictions and the ground-truth optical flows. $\gamma$ is the optimizing noise level and $k$ is the skip length, introduced in Section \ref{sec:accelerate}.}
  \label{tab:accelerate}
  \centering
  \begin{tabular}{lccccl}
    \toprule
    & \multicolumn{2}{c}{Sintel \cite{butler2012sintel}} & \multicolumn{2}{c}{VKITTI \cite{gaidon2016vkitti}} & \multirow{2}{*}{Average speed-up} \\
    \cmidrule(r){2-3}\cmidrule(r){4-5}
    $\gamma$ & Depth & Normal & Depth & Normal \\
    \midrule
    2  & 10.13 &  8.51 & 22.73 & 23.97 & 3.07$\times$
        \begin{tikzpicture}[scale=1.5]
        \filldraw[blue!61.4!lightgray] (0,0) rectangle (3.07,.15);
        \end{tikzpicture} \\
    4  & 10.93 &  9.46 & 22.75 & 23.88 & 1.99$\times$
        \begin{tikzpicture}[scale=1.5]
        \filldraw[blue!39.8!lightgray] (0,0) rectangle (1.99,.15);
        \end{tikzpicture} \\
    7  & 12.83 & 10.95 & 22.18 & 23.52 & 1.31$\times$
        \begin{tikzpicture}[scale=1.5]
        \filldraw[blue!26.2!lightgray] (0,0) rectangle (1.31,.15);
        \end{tikzpicture} \\
    10 & 14.21 & 12.01 & 21.38 & 22.86 & 1.00$\times$
        \begin{tikzpicture}[scale=1.5]
        \filldraw[blue!20!lightgray] (0,0) rectangle (1,.15);
        \end{tikzpicture} \\
    \midrule
    $k$ \\
    \midrule
    4 & 12.87 & 11.99 & 20.34 & 21.27 & 3.66$\times$
        \begin{tikzpicture}[scale=1.5]
        \filldraw[blue!73.2!lightgray] (0,0) rectangle (3.66,.15);
        \end{tikzpicture} \\
    3 & 13.33 & 11.86 & 20.40 & 21.44 & 2.84$\times$
        \begin{tikzpicture}[scale=1.5]
        \filldraw[blue!57!lightgray] (0,0) rectangle (2.85,.15);
        \end{tikzpicture} \\
    2 & 14.06 & 11.70 & 20.66 & 21.78 & 1.97$\times$
        \begin{tikzpicture}[scale=1.5]
        \filldraw[blue!21.4!lightgray] (0,0) rectangle (1.07,.15);
        \end{tikzpicture} \\
    1 & 14.21 & 12.01 & 21.38 & 22.86 & 1.00$\times$
        \begin{tikzpicture}[scale=1.5]
        \filldraw[blue!20!lightgray] (0,0) rectangle (1,.15);
        \end{tikzpicture} \\
    \bottomrule
  \end{tabular}
\end{table}

\section{Related work}
\subsection{Image-to-image translation}
Isola \etal \cite{isola2017pix2pix} (Pix2Pix) introduced a supervised framework for mapping images using conditional generative neural networks \cite{mirza2014conditional}. Building upon Pix2Pix, numerous works \cite{park2019semantic,park2019semantic,liu2019learning,li2019linestofacephoto} have extended the original idea to various image translation tasks. Although they are able to capture the relationship between the input layout and the desired output image, these methods rely on paired data during training. With the success of large text-to-image models \cite{rombach2022LDM,saharia2022imagen,ramesh2022dalle2}, Zhang and Agrawala \cite{zhang2023controlnet} leverage these models to perform text-driven image translation to break the limitation of predefined domains.

\subsection{Consistent image translation}
While the aforementioned image translation methods can be directly extended to videos by treating each frame independently, the omission of temporal consistency can result in the presence of flicker artifacts in the generated videos. To mitigate flicker artifacts and ensure temporal consistency, Ruder \etal \cite{ruder2016artistic} leverage optical flow information to warp the previously generated frame onto the current frame, which aids in frame initialization and prevents the optimization process from getting trapped in local minima. In a similar vein, Huang \etal \cite{huang2017real} introduced a feed-forward network that utilizes a similar temporal loss to ensure both stylization and temporal consistency in consecutive frames. By training the network with this objective, they aim to generate outputs that exhibit consistent stylization across frames while maintaining temporal coherence throughout the video sequence.

In contrast to their approach, our method performs optimization by tuning the latent samples, unlocking the full potential of the pre-trained text-to-image models. Furthermore, our approach does not rely on a specific style example image to provide style reference. Instead, Video ControlNet allows more flexibility by incorporating text-driven style customization and can perform synthetic-to-real video translation.

\subsection{Diffusion model for video generation}

Meanwhile, several recent works have extended diffusion models to the domain of video generation \cite{harvey2022flexible,ho2022video,hoppe2022diffusion,voleti2022masked}. However, these methods often come with certain limitations. They typically require additional training data and longer training processes, which can be challenging due to data availability and the computational overhead involved. In contrast, our approach offers a more convenient and efficient video generation. It eliminates the need for extensive additional training data and significantly reduces the training time required for generating video. This convenience becomes particularly valuable when the goal is to efficiently produce videos in higher resolutions. Furthermore, our approach is compatible with existing large pre-trained diffusion models, which helps overcome computational limitations that may arise during the generation process.



\section{Conclusion}
In this work, we show how to exploit an off-the-shelf conditional image DM for temporally consistent synthetic-to-real video translation. With the correspondence information from the optical flow of the synthetic videos, we can formulate a constrained optimization problem to enforce the temporal consistency constraint and to find the optimal initial noisy latents for temporal coherent video synthesis using DMs. In addition, it is worth noting that the proposed method does not involve any further training or fine-tuning of the diffusion models. With extensive experiments on various benchmarks for synthetic-to-real video translation, the results demonstrate the superior performance of our approach over compared baseline methods, both quantitatively and qualitatively.

\paragraph{Negative Societal Impact} We are aware of the fact that our method may boost the possibility to generate realistic and convincing videos that can be indistinguishable from genuine footage. This raises concerns about the potential for AI-generated videos to be used for spreading false information, creating fake news, or manipulating public opinion. For this reason, we are not planning to release the full implementation of Video ControlNet to the public until robust detection and verification systems that can identify AI-generated videos are maturely developed. However, Video ControlNet also have positive influences on, for example, robotic learning and graphics rendering. We will work closely with teams that can derive significant benefits from our work, under a rigorous protocol.

\bibliographystyle{plain}
\bibliography{refs}

\begin{thebibliography}{10}

\bibitem{bernardin2014clear}
Keni Bernardin and Rainer Stiefelhagen.
\newblock Evaluating multiple object tracking performance: The clear mot
  metrics.
\newblock {\em EURASIP Journal on Image and Video Processing}, 2008.

\bibitem{brooks2023instructpix2pix}
Tim Brooks, Aleksander Holynski, and Alexei~A. Efros.
\newblock Instructpix2pix: Learning to follow image editing instructions.
\newblock In {\em Proceedings of the IEEE/CVF Conference on Computer Vision and
  Pattern Recognition}, 2023.

\bibitem{butler2012sintel}
D.~J. Butler, J.~Wulff, G.~B. Stanley, and M.~J. Black.
\newblock A naturalistic open source movie for optical flow evaluation.
\newblock In {\em Proceeding of the European Conference on Computer Vision},
  pages 611--625, 2012.

\bibitem{blender}
Blender~Online Community.
\newblock {\em Blender - a 3D modelling and rendering package}.
\newblock Blender Foundation, Stichting Blender Foundation, Amsterdam, 2023.

\bibitem{devlin2019bert}
Jacob Devlin, Ming-Wei Chang, Kenton Lee, and Kristina Toutanova.
\newblock {BERT}: Pre-training of deep bidirectional transformers for language
  understanding.
\newblock In {\em Proceedings of the 2019 Conference of the North {A}merican
  Chapter of the Association for Computational Linguistics: Human Language
  Technologies, Volume 1 (Long and Short Papers)}, pages 4171--4186, 2019.

\bibitem{dhariwal2021diffusion}
Prafulla Dhariwal and Alexander Nichol.
\newblock Diffusion models beat gans on image synthesis.
\newblock In {\em Advances in Neural Information Processing Systems}, pages
  8780--8794, 2021.

\bibitem{gaidon2016vkitti}
A~Gaidon, Q~Wang, Y~Cabon, and E~Vig.
\newblock Virtual worlds as proxy for multi-object tracking analysis.
\newblock In {\em Proceedings of the IEEE/CVF Conference on Computer Vision and
  Pattern Recognition}, 2016.

\bibitem{harvey2022flexible}
William Harvey, Saeid Naderiparizi, Vaden Masrani, Christian Weilbach, and
  Frank Wood.
\newblock Flexible diffusion modeling of long videos.
\newblock In {\em Advances in Neural Information Processing Systems}, pages
  27953--27965, 2022.

\bibitem{ho2020ddpm}
Jonathan Ho, Ajay Jain, and Pieter Abbeel.
\newblock Denoising diffusion probabilistic models.
\newblock In {\em Advances in Neural Information Processing Systems}, 2020.

\bibitem{ho2022video}
Jonathan Ho, Tim Salimans, Alexey Gritsenko, William Chan, Mohammad Norouzi,
  and David~J Fleet.
\newblock Video diffusion models.
\newblock In {\em Advances in Neural Information Processing Systems}, pages
  8633--8646, 2022.

\bibitem{hoppe2022diffusion}
Tobias H{\"o}ppe, Arash Mehrjou, Stefan Bauer, Didrik Nielsen, and Andrea
  Dittadi.
\newblock Diffusion models for video prediction and infilling.
\newblock {\em Transactions on Machine Learning Research}, 2022.

\bibitem{huang2017real}
Haozhi Huang, Hao Wang, Wenhan Luo, Lin Ma, Wenhao Jiang, Xiaolong Zhu, Zhifeng
  Li, and Wei Liu.
\newblock Real-time neural style transfer for videos.
\newblock In {\em Proceedings of the IEEE Conference on Computer Vision and
  Pattern Recognition}, pages 783--791, 2017.

\bibitem{isola2017pix2pix}
Phillip Isola, Jun-Yan Zhu, Tinghui Zhou, and Alexei~A Efros.
\newblock Image-to-image translation with conditional adversarial networks.
\newblock In {\em Proceedings of the IEEE/CVF Conference on Computer Vision and
  Pattern Recognition}, pages 1125--1134, 2017.

\bibitem{kingma2013adam}
Diederik~P Kingma and Jimmy Ba.
\newblock Adam: A method for stochastic optimization.
\newblock {\em arXiv preprint arXiv:1412.6980}, 2014.

\bibitem{kingma14vae}
Diederik~P Kingma and Max Welling.
\newblock Auto-encoding variational bayes.
\newblock In {\em International Conference on Learning Representations}, 2014.

\bibitem{li2019linestofacephoto}
Yuhang Li, Xuejin Chen, Feng Wu, and Zheng-Jun Zha.
\newblock Linestofacephoto: Face photo generation from lines with conditional
  self-attention generative adversarial networks.
\newblock In {\em ACM Multimedia}, pages 2323--2331, 2019.

\bibitem{liu2019learning}
Xihui Liu, Guojun Yin, Jing Shao, Xiaogang Wang, et~al.
\newblock Learning to predict layout-to-image conditional convolutions for
  semantic image synthesis.
\newblock {\em Advances in Neural Information Processing Systems}, 32, 2019.

\bibitem{mirza2014conditional}
Mehdi Mirza and Simon Osindero.
\newblock Conditional generative adversarial nets.
\newblock {\em arXiv preprint arXiv:1411.1784}, 2014.

\bibitem{nichol2021improved}
Alexander~Quinn Nichol and Prafulla Dhariwal.
\newblock Improved denoising diffusion probabilistic models.
\newblock In {\em International Conference on Machine Learning}, pages
  8162--8171, 2021.

\bibitem{park2019semantic}
Taesung Park, Ming-Yu Liu, Ting-Chun Wang, and Jun-Yan Zhu.
\newblock Semantic image synthesis with spatially-adaptive normalization.
\newblock In {\em Proceedings of the IEEE/CVF Conference on Computer Vision and
  Pattern Recognition}, pages 2337--2346, 2019.

\bibitem{radford2021clip}
Alec Radford, Jong~Wook Kim, Chris Hallacy, Aditya Ramesh, Gabriel Goh,
  Sandhini Agarwal, Girish Sastry, Amanda Askell, Pamela Mishkin, Jack Clark,
  Gretchen Krueger, and Ilya Sutskever.
\newblock Learning transferable visual models from natural language
  supervision.
\newblock In {\em International Conference on Machine Learning}, pages
  8748--8763, 2021.

\bibitem{ramesh2022dalle2}
Aditya Ramesh, Prafulla Dhariwal, Alex Nichol, Casey Chu, and Mark Chen.
\newblock Hierarchical text-conditional image generation with clip latents,
  2022.

\bibitem{rombach2022LDM}
Robin Rombach, Andreas Blattmann, Dominik Lorenz, Patrick Esser, and Bj\"orn
  Ommer.
\newblock High-resolution image synthesis with latent diffusion models.
\newblock In {\em Proceedings of the IEEE/CVF Conference on Computer Vision and
  Pattern Recognition}, pages 10684--10695, 2022.

\bibitem{ruder2016artistic}
Manuel Ruder, Alexey Dosovitskiy, and Thomas Brox.
\newblock Artistic style transfer for videos.
\newblock In {\em German Conference on Pattern Recognition}, pages 26--36,
  2016.

\bibitem{ruiz2022dreambooth}
Nataniel Ruiz, Yuanzhen Li, Varun Jampani, Yael Pritch, Michael Rubinstein, and
  Kfir Aberman.
\newblock Dreambooth: Fine tuning text-to-image diffusion models for
  subject-driven generation.
\newblock {\em arXiv preprint arxiv:2208.12242}, 2022.

\bibitem{saharia2022palette}
Chitwan Saharia, William Chan, Huiwen Chang, Chris Lee, Jonathan Ho, Tim
  Salimans, David Fleet, and Mohammad Norouzi.
\newblock Palette: Image-to-image diffusion models.
\newblock In {\em ACM SIGGRAPH}, 2022.

\bibitem{saharia2022imagen}
Chitwan Saharia, William Chan, Saurabh Saxena, Lala Li, Jay Whang, Emily~L
  Denton, Kamyar Ghasemipour, Raphael Gontijo~Lopes, Burcu Karagol~Ayan, Tim
  Salimans, Jonathan Ho, David~J Fleet, and Mohammad Norouzi.
\newblock Photorealistic text-to-image diffusion models with deep language
  understanding.
\newblock In {\em Advances in Neural Information Processing Systems}, pages
  36479--36494, 2022.

\bibitem{schuhmann2022laion5b}
Christoph Schuhmann, Romain Beaumont, Richard Vencu, Cade Gordon, Ross
  Wightman, Mehdi Cherti, Theo Coombes, Aarush Katta, Clayton Mullis, Mitchell
  Wortsman, Patrick Schramowski, Srivatsa Kundurthy, Katherine Crowson, Ludwig
  Schmidt, Robert Kaczmarczyk, and Jenia Jitsev.
\newblock Laion-5b: An open large-scale dataset for training next generation
  image-text models, 2022.

\bibitem{shoemake1985slerp}
Ken Shoemake.
\newblock Animating rotation with quaternion curves.
\newblock In {\em Proceedings of the 12th annual conference on Computer
  graphics and interactive techniques}, page 245–254, 1985.

\bibitem{song2021ddim}
Jiaming Song, Chenlin Meng, and Stefano Ermon.
\newblock Denoising diffusion implicit models.
\newblock In {\em International Conference on Learning Representations}, 2021.

\bibitem{song2021score}
Yang Song, Jascha Sohl{-}Dickstein, Diederik~P. Kingma, Abhishek Kumar, Stefano
  Ermon, and Ben Poole.
\newblock Score-based generative modeling through stochastic differential
  equations.
\newblock In {\em International Conference on Learning Representations}, 2021.

\bibitem{vaswani2017transformer}
Ashish Vaswani, Noam Shazeer, Niki Parmar, Jakob Uszkoreit, Llion Jones,
  Aidan~N Gomez, \L~ukasz Kaiser, and Illia Polosukhin.
\newblock Attention is all you need.
\newblock In {\em Advances in Neural Information Processing Systems}, 2017.

\bibitem{voleti2022masked}
Vikram Voleti, Alexia Jolicoeur-Martineau, and Christopher Pal.
\newblock {MCVD} - masked conditional video diffusion for prediction,
  generation, and interpolation.
\newblock In {\em Advances in Neural Information Processing Systems}, 2022.

\bibitem{xu2022gmflow}
Haofei Xu, Jing Zhang, Jianfei Cai, Hamid Rezatofighi, and Dacheng Tao.
\newblock Gmflow: Learning optical flow via global matching.
\newblock In {\em Proceedings of the IEEE/CVF Conference on Computer Vision and
  Pattern Recognition}, pages 8121--8130, 2022.

\bibitem{zhang2023controlnet}
Lvmin Zhang and Maneesh Agrawala.
\newblock Adding conditional control to text-to-image diffusion models.
\newblock {\em arXiv preprint arxiv:2302.05543}, 2023.

\bibitem{zhou2020centertrack}
Xingyi Zhou, Vladlen Koltun, and Philipp Kr{\"a}henb{\"u}hl.
\newblock Tracking objects as points.
\newblock In {\em Proceeding of the European Conference on Computer Vision},
  pages 474--490, 2020.

\end{thebibliography}


\newpage
\appendix

\section{Experiment details}
This section gives details on experimental setup, architectures and optimization hyperparameters. In addition, comparative videos that visually demonstrate the superiority of our method can be found in the accompanying media files.

\subsection{Diffusion settings}
We use the publicly available implementation and weights of ControlNet from Hugging face Diffusers\footnote{\url{https://github.com/huggingface/diffusers}}, which accepts arbitrary visual input/output dimensions as long as the width and height are multiples of eight. The weight for backbone SD is downloaded from \url{https://huggingface.co/runwayml/stable-diffusion-v1-5}. The ControlNet adapters for depth conditioning and normal conditioning are downloaded from \url{https://huggingface.co/lllyasviel/control_v11f1p_sd15_depth} and \url{https://huggingface.co/lllyasviel/control_v11p_sd15_normalbae}, respsctively. The classifier-free guidance scale is set to 7.5, and ControlNet conditioning scale is set to 1 for all experiments. For scheduler, we use DDIM with default configs.

\subsection{Datasets}
\paragraph{Occlusion derivation}
We derive the occlusion maps from the optical flows for videos in VKITTI \cite{gaidon2016vkitti}. Concretely, we use the optical flows to find the correspondence of the accompanying rendered animation voxels. If the discrepancy in voxel value of adjacent frames surpasses a given threshold, or the correspondence is out-of-frame, we mark the voxel coordinate as occluded.

\paragraph{Normal derivation}
We derive the per-pixel surface normal from the partial derivatives of the depth map. The derivatives are adjust by the camera intrinsic matrix exported from the 3D modeling software and normalized by the depth. We provide a concrete implementation to encourage reproducibility.
\begin{lstlisting}
def depth2normal(depths, cams):
    """
    Calculate normal from depth
    args:
        depths: float tensor in [T, 1, H, W] format
        cams: camera intrinsic, float tensor in [T, 3, 3] format
    """
    dx, dy = torch.gradient(depths, dim=[2, 3])
    dx *= cams[:, 0, 0].view(-1, 1, 1, 1) / depths
    dy *= cams[:, 1, 1].view(-1, 1, 1, 1) / depths
    normal = torch.cat((-dx, -dy, torch.ones_like(depths)), dim=1)
    normal /= ((normal ** 2).sum(dim=1, keepdims=True) ** 0.5)
    normal = (normal * 0.5 + 0.5).clip(0, 1)
    return normal
\end{lstlisting}



\section{Implementation details}
\label{appendix:implementation}
In this section, we present how the proposed Video ControlNet as introduced in Section \ref{sec:video-controlnet} can be implemented in Pytorch\footnote{\url{https://pytorch.org}} to help the readers better comprehend the whole algorithm.  Our proposed algorithm consists of two major modules: (1) flow warping and masking, and (2) memory-efficient temporal consistency optimization. Their details are shown as follows.


\subsection{Flow warping and masking}
We implement the proposed flow warping, Eq. \eqref{eq:flow-warping}, and masking, Eq. \eqref{eq:masking}, in a simple nearest-neighbor interpolation scheme, while more sophisticated interpolation methods, such as bilinear and bicubic interpolations, can also be applied with little modifications 
to the code snippet below.
\begin{lstlisting} 
def flow_warping(frame, flow, occlusion):
    """
    Warp (t)th frame to (t-1)th frame
    args:
        frame: float tensor in [H, W, 3] format
        flow: float tensor in [H, W, 2] format
        occlusion: bool tensor in [H, W] format
    """
    # create a matrix where each entry is its coordinate
    s = flows.shape[:2]
    m = np.array([i for i in np.ndindex(*s)])
    meshgrid = torch.from_numpy(m.reshape(*s, len(s)))

    # find the coordinates that the flows point to
    dest = flow + meshgrid
    
    # discard out-of-frame flows
    valid_mask = ((dest >= 0) &
                  (dest < torch.tensor(dest.shape[:2]) - 1)).all(-1)
    v_src = meshgrid[valid_mask]
    
    # nearest-neighbor warping
    v_dst = dest[valid_mask]
    v_dst = v_dst.round().to(int)

    # warp by flow
    valid_pixels = frame[v_dst[:, 0], v_dst[:, 1]]
    warped[v_src[:, 0], v_src[:, 1]] = valid_pixels

    # mask occluded pixels
    occlusion_coord = meshgrid[occlusion]
    warped[occlusion_coord[:, 0], occlusion_coord[:, 1]] = 0

    return warped
\end{lstlisting}
As shown in the pseudo-code, the pixel correspondence in the flow warping algorithm is invariant to the pixel values of the input frame. We can therefore pre-compute the correspondence and reuse it for different frames.

\subsection{Memory-efficient temporal consistency optimization}
\begin{figure}
    \centering
    \vspace{3cm}
    \includegraphics[width=1\linewidth]{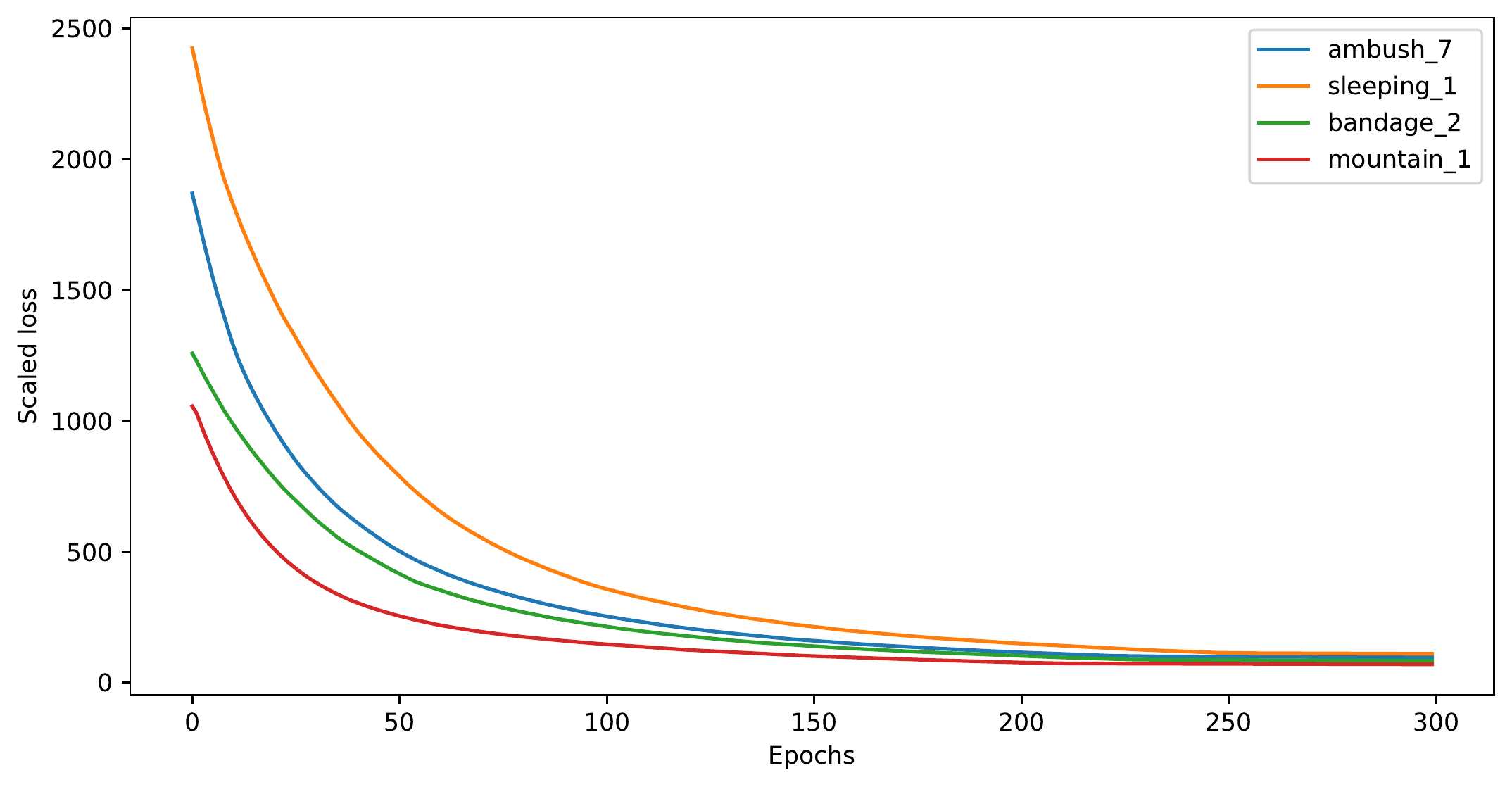}
    \caption{The memory-efficient implementation of the temporal consistency optimization helps a steady optimizing process. The loss converges around 150 epochs. The loss is scaled by the GradeScaler in Pytorch$^\dagger$.}
    \raggedright
    \small$^\dagger$ \url{https://pytorch.org/docs/stable/amp#torch.cuda.amp.GradScaler}
    \label{fig:loss}
    \vspace{2cm}
\end{figure}

While the optimal noise $z^L_*$ in Eq. \eqref{eq:opt-noise} can be crafted by Gradient Descent, forward and backward propagating the entire video at once can pose challenges in terms of memory consumption and scalability for long videos. To address this issue, we leverage the gradient accumulation in Pytorch and propagate one frame at a time. Concretely, we keep a cache list in CPU for the generated frames which are detached from their computation tree. When frame $I_t$ is sampled from the DM, $I_t$ is warped back and compared with the cached frames to calculate the loss. The gradients w.r.t. the loss are then backpropagated to $z^L_t$, and the computation graph for $I_t$ can now be freed. In such implementation, only one computation graph for a frame stays alive at a time, substantially reducing the memory requirement. However, $z^L_0$ never receives gradient in this setting, so we specifically allow the cache frames from last iteration to be warped back and compared with $I_0$ when it computation graph is built, as shown in the pseudo-code below. As shown in Figure \ref{fig:loss}, our implementation produces healthy and predictable loss curves during the optimization.

\newpage
\begin{lstlisting}
def temporal_consistency_optimization(
    noise, cond, prompt, n_epochs, flows, occlusions):
    """
    Craft an optimal noise to generate temporally consistent video
    args:
        noise: float tensor in [T, H, W, C] format
        cond: float tensor in [T, H, W, C] format
        prompt: textual conditioning string
        n_epochs: number of epochs of the optimization
        flow: float tensor in [T-1, H, W, 2] format
        occlusion: bool tensor in [T-1, H, W] format
    """
    optimizer = SGD(noise)
    previous_frames = None
    for _ in range(n_epochs):
        optimizer.zero_grad()
        for i in range(-1, len(flows)):
            frame = ControlNet(
                noise[i+1:i+2], cond[i+1:i+2].cuda(), prompt)
            ######################################################
            # the 0th frame is handled differently
            if i == -1:
                if previous_frames:
                    loss = []
                    for j in range(min(window_size,
                                       len(previous_frames) - 1)):
                        # warp jth frame (from last iter) to
                        # the 0th position
                        warped = previous_frames[j+1].cuda()
                        while j >= 0:
                            warped = flow_warping(warped, flows[j],
                                                  occlusions[j])
                            j -= 1
                        err = torch.where(warped != 0,
                                          warped - frame, 0) ** 2
                        # normalized by number of non-zero pixels
                        loss.append(err.sum() / (err!=0).sum())
                    loss = sum(loss)/ len(loss)
                    loss.backward()
                previous_frames = [frame.detach().cpu()]
                continue
            ######################################################
            previous_frames.append(frame.detach().cpu())
            warped = frame
            loss = []
            for offset in range(window_size):
                j = i - offset
                if j < 0:
                    break # we reach beyond the start of the video
                ref = previous_frames[j].cuda()
                warped = flow_warping(warped, flows[j],
                                      occlusions[j])
                err = torch.where(warped != 0, warped - ref, 0) ** 2
                # normalized by number of non-zero pixels
                loss.append(err.sum() / (err!=0).sum())
            loss = sum(loss)/ len(loss)
            loss.backward()
        optimizer.step()
\end{lstlisting}

\end{document}